\documentclass[letterpaper]{article} 
\usepackage{aaai23}  
\usepackage{times}  
\usepackage{helvet}  
\usepackage{courier}  
\usepackage[hyphens]{url}  
\usepackage{graphicx} 
\usepackage{natbib}  
\usepackage{caption} 
\usepackage{algorithm}
\usepackage{algorithmic}

\usepackage{newfloat}
\usepackage{listings}
\DeclareCaptionStyle{ruled}{labelfont=normalfont,labelsep=colon,strut=off} 
\floatstyle{ruled}
\newfloat{listing}{tb}{lst}{}
\floatname{listing}{Listing}
\newif\ifappendix
\appendixtrue 

\usepackage{amsmath,amsthm,amssymb}
\usepackage{mathtools}
\usepackage{bm}
\usepackage{dsfont}
\usepackage{xspace}
\usepackage{color,soul}
\usepackage{multirow}
\usepackage{multicol}
\usepackage{caption}
\usepackage{enumitem}

\newcommand{\sys}{ProxyBO\xspace}

\newtheorem{theorem}{Theorem}

\newtheorem*{assumption*}{\assumptionnumber}
\providecommand{\assumptionnumber}{}
\makeatletter
\newenvironment{assumption}[1]
 {%
  \renewcommand{\assumptionnumber}{Assumption #1}%
  \begin{assumption*}%
  \protected@edef\@currentlabel{#1}%
 }
 {%
  \end{assumption*}
 }
\makeatother

\newtheorem*{remark*}{\remarknumber}
\providecommand{\remarknumber}{}


\usepackage{microtype}
\usepackage{graphicx}
\usepackage{subfigure}
\usepackage{booktabs} 

\usepackage{algorithm}
\usepackage{algorithmic}
\renewcommand{\algorithmicrequire}{\textbf{Input:}}
\renewcommand{\algorithmicensure}{\textbf{Output:}}

\title{\sys: Accelerating Neural Architecture Search via \\ Bayesian Optimization with Zero-Cost Proxies}
\author{
    Yu Shen\textsuperscript{\rm 1,4},
    Yang Li\textsuperscript{\rm 5},
    Jian Zheng\textsuperscript{\rm 3},
    Wentao Zhang\textsuperscript{\rm 6,7},
    Peng Yao\textsuperscript{\rm 4},\\
    Jixiang Li\textsuperscript{\rm 4},
    Sen Yang\textsuperscript{\rm 4},
    Ji Liu\textsuperscript{\rm 4},
    Bin Cui\textsuperscript{\rm 1,2}
    \\
}

\affiliations{
    \textsuperscript{\rm 1}Key Lab of High Confidence Software Technologies, Peking University, China\\
    \textsuperscript{\rm 2}Institute of Computational Social Science, Peking University (Qingdao), China\\
    \textsuperscript{\rm 3}School of Computer Science and Engineering, Beihang University, China\\
    \textsuperscript{\rm 4}Kuaishou Technology, China\\
    \textsuperscript{\rm 5}Data Platform, TEG, Tencent Inc., China\\
    \textsuperscript{\rm 6}Mila - Québec AI Institute\\
    \textsuperscript{\rm 7}HEC, Montréal, Canada\\
    
    \{shenyu,bin.cui\}@pku.edu.cn,   
    zhengjian2322@buaa.edu.cn,\\
    \{yaopeng,lijixiang,senyang\}@kuaishou.com, jiliu@kwai.com, \\
    thomasyngli@tencent.com, wentao.zhang@mila.quebec\\
}
\begin{document}
\maketitle

\begin{abstract}
Designing neural architectures requires immense manual efforts. 
This has promoted the development of neural architecture search (NAS) to automate the design. 
While previous NAS methods achieve promising results but run slowly, zero-cost proxies run extremely fast but are less promising.
Therefore, it's of great potential to accelerate NAS via those zero-cost proxies.
The existing method has two limitations, which are \textit{unforeseeable reliability} and \textit{one-shot usage}.
To address the limitations, we present \sys, an efficient Bayesian optimization (BO) framework that utilizes the zero-cost proxies to accelerate neural architecture search. 
We apply the generalization ability measurement to estimate the fitness of proxies on the task during each iteration and design a novel acquisition function to combine BO with zero-cost proxies based on their dynamic influence.
Extensive empirical studies show that \sys consistently outperforms competitive baselines on five tasks from three public benchmarks. 
Concretely, \sys achieves up to $5.41\times$ and $3.86\times$ speedups over the state-of-the-art approaches REA and BRP-NAS.
\end{abstract}

\section{Introduction}
\label{sec:introduction}
Discovering state-of-the-art neural architectures ~\cite{he2016deep,huang2017densely} requires substantial efforts of human experts. 
The manual design is often costly and becomes increasingly expensive when networks grow larger. 
Recently, the neural network community has witnessed the development of neural architecture search (NAS)~\cite{zoph2018learning,real2019regularized,liu2018darts,cai2018efficient}, which turns the design of architectures into an optimization problem without human interaction and achieves promising results in a wide range of fields, such as 
image classification~\cite{zoph2018learning,real2019regularized}, 
sequence modeling~\cite{pham2018efficient,so2019evolved}, etc.   

Bayesian optimization (BO)~\cite{hutter2011sequential,snoek2012practical} has emerged as a state-of-the-art method for NAS~\cite{ying2019bench,white2021bananas}. It trains a predictor, namely surrogate, on observations and selects the next architecture to evaluate based on its predictions. Recent work differs in the choice of surrogate, including Bayesian neural networks~\cite{springenberg2016bayesian}, Graph neural networks~\cite{ma2019deep}, etc. 
Despite the promising converged results, they share the common drawback that training a well-performed surrogate requires a sufficient number of evaluations, which often take days to obtain.

While several approaches attempt to reduce the evaluation cost via weight sharing~\cite{pham2018efficient} or gradient decent~\cite{liu2018darts}, 
recent work~\cite{chen2020neural,mellor2021neural} proposes several zero-cost proxies to estimate the performance of architecture at initialization using only a few seconds instead of network training.
Though they achieve less promising results than BO-based methods, the computation of zero-cost proxies is speedy. 
Then, there comes up a question:
``\textit{Can we speed up NAS by combining the advantages of both Bayesian optimization and zero-cost proxies, i.e., achieving promising results with fewer computationally expensive evaluations?}''

\begin{figure}
    \centering
    \subfigure[NAS-Bench-201 CIFAR-10]{
		\scalebox{0.46}{
			\includegraphics[width=1\linewidth]{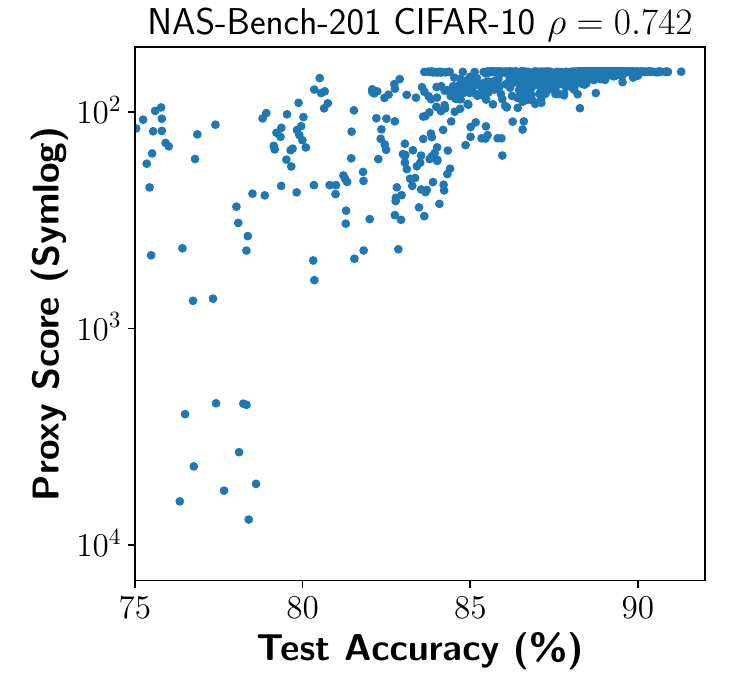}
			\label{fig:intro_201}
	}}
	\subfigure[NAS-Bench-101]{
		\scalebox{0.46}{
			\includegraphics[width=1\linewidth]{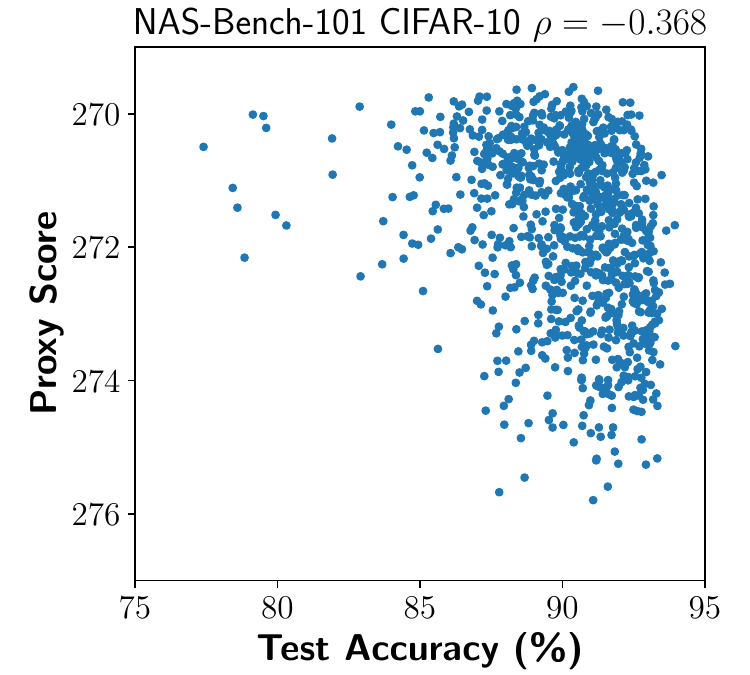}
			\label{fig:intro_101}
	}}
    \caption{Spearman $\rho$ of \texttt{jacob\_cov} over NAS search spaces using 1000 randomly sampled architectures.}
    \label{fig:intro}
\end{figure}


\noindent\textbf{Opportunities.}
As shown in Figure~\ref{fig:intro_201}, the Spearman correlation coefficient between the proxy \texttt{jacob\_cov} and test accuracy on NAS-Bench-201 CIFAR-10 is 0.742.
Since the coefficient is a relatively large positive value, the proxy can be applied to rank architectures and guide the selection of the next architecture to evaluate.  

\noindent\textbf{Challenges.}
First, utilizing zero-cost proxies is non-trivial. 
Rather than applying Bayesian optimization, recent work~\cite{abdelfattah2020zero} attempts to perform a simple warm-up on binary relation predictors based on a specific proxy.
However, the proxies are not fully utilized via warm-up and may even lead to negative effects due to two limitations: 
\textbf{\textit{L1. Unforeseeable reliability}}.
The warm-up method chooses to apply the best proxy found by exhaustively evaluating thousands of architectures over a specific search space. 
However, in practice, the correlation between proxy scores and objective values is unknown before searching. 
As shown in Figure~\ref{fig:intro}, although the proxy \texttt{jacob\_cov} works well on NAS-Bench-201, it performs worse than ranking randomly on NAS-Bench-101. In this case, using this proxy may lead to negative effects; 
\textbf{\textit{L2. One-shot usage}}.
The warm-up method applies the proxies only once before searching by pre-training the neural binary predictor based on proxy scores.
As a result, the influence of proxies in the warm-up method is irreversible, making it difficult to identify and get rid of those potentially bad proxies during searching.
Due to the limitations, how to unleash the potential of the proxies is still an open question.

In addition, though BO and zero-cost proxies may complement with each other, combining the two parts is also non-trivial. 
As a learning model, the BO surrogate generalizes better with more evaluation results while the ranking ability of proxies is constant, which is only determined by the current task. 
In other words, zero-cost proxies bring benefits when few evaluations are given, but they become less helpful when the BO surrogate becomes accurate with sufficient evaluations.
Therefore, the influence of the proxies should be decreased during optimization, and a dynamic design should be made to match this trend.

\noindent\textbf{Contributions.} 
In this paper, we propose \sys, an efficient Bayesian optimization (BO) framework that utilizes the zero-cost proxies to significantly accelerate neural architecture search. 
The contributions are summarized as follows:
1) To the best of our knowledge, \sys is the first method that utilizes the zero-cost proxies without prior knowledge about whether the proxies are suitable for the current task; 
2) To deal with \textbf{L1}, we propose to estimate the reliability of different proxies by measuring their fitness during optimization. 
For \textbf{L2}, we apply a novel acquisition function to combine BO with the proxies based on their dynamic influence;
3) Empirical results on four public benchmarks showcase the superiority of \sys compared with \textbf{fifteen} competitive baselines. Concretely, \sys achieves up to $5.41\times$ and $3.86\times$ speedups over the state-of-the-art approaches REA and BRP-NAS, respectively.


\section{Related Work}
Designing neural architectures manually is often a challenging and time-consuming task since it is quite difficult for human experts to choose the proper operations and place each connection. 
As a result, this arouses great interest from both academia and industry to design neural architectures in an automatic manner. 
Inspired by successful hand-crafted architectures, pioneering work~\cite{zoph2018learning} manually designs a fixed macro-structure, which is composed of stacks of cells (micro-structure). 
The search is then conducted over those cells instead of the whole architecture via different search strategies, including reinforcement learning~\cite{zoph2018learning,pham2018efficient}, evolutionary algorithm~\cite{real2019regularized}, gradient decent~\cite{liu2018darts}, binary relation predictor~\cite{dudziak2020brp}, etc. 

Among various approaches proposed for neural architecture search, recent researches~\cite{white2021bananas,ying2019bench,siems2020bench} have shown the competitive performance of Bayesian optimization (BO) with a performance predictor.
The original BO~\cite{snoek2012practical} is proposed for solving black-box optimization, in which the output can only be obtained by an objective function, and no extra information like derivatives is available. 
As evaluating the validation performance of a given neural architecture is also a black-box process, BO can be directly applied to search for neural architectures.
Benchmark studies~\cite{ying2019bench,siems2020bench} point out that SMAC~\cite{hutter2011sequential}, a classical BO approach, achieves state-of-the-art performance given enough budgets. 
BANANAS~\cite{white2021bananas} digs deeper into the BO framework and compares each part of the framework via extensive experiments.
Other work further improves BO by combining the characteristics of neural architectures, e.g., NASBOT~\cite{kandasamy2018neural} defines a pseudo-distance for kernel functions while GPWL~\cite{ru2020interpretable} adopts the Weisfeiler-Lehman kernel. 
However, these methods share the same drawback that a sufficient number of evaluations are required to guide the BO framework. As the evaluation of architectures is time-consuming, the cost of the search algorithm is exorbitant.

To reduce the costs of NAS, several techniques have been applied in the literature. 
ENAS~\cite{pham2018efficient} applies the weight sharing strategy by sharing weights in the same operation.
DARTS~\cite{liu2018darts} models NAS as training an over-parameterized architecture including all candidate paths.
EcoNAS~\cite{zhou2020econas} investigates proxies with reduced resources during evaluation, e.g., fewer epochs, training samples, etc. 
BOHB~\cite{falkner2018bohb}
combines Hyperband~\cite{li2017hyperband} and BO by early-stopping bad evaluations while MFES~\cite{li2021mfes,li2022hyper} improves BOHB by taking all evaluations of different resource levels into consideration when sampling new configurations to evaluate.
Transfer learning~\cite{lee2021rapid} are applied to learn from previous tasks.

Recent studies further accelerate NAS by estimating the performance of architectures at initialization (i.e., zero-cost proxies~\cite{xu2021knas,shu2021nasi}). 
Synaptic saliency metrics~\cite{lee2018snip,wang2019picking,tanaka2020pruning} measures the loss change when removing a certain parameter.
Fisher~\cite{theis2018faster} estimates the loss change when removing activation channels.
NASWOT~\cite{mellor2021neural} and TE-NAS~\cite{chen2020neural} model the expressivity of architectures based on activations. 
HNAS~\cite{shu2022unifying} boosts training-free NAS in a principled way.
As proxy scores are somehow related to the ground-truth performance, recent work~\cite{abdelfattah2020zero} proposes to warm up a neural binary relation predictor or select candidates for evolutionary algorithms using a specific proxy and achieves acceleration on benchmarks. 
OMNI~\cite{white2021powerful} combines proxies into the BO surrogate.
Due to the limitations (\textbf{L1} and \textbf{L2}), how to unleash the potential of zero-cost proxies without prior knowledge is still an open question.

\section{Preliminary}
As stated above, neural architecture search (NAS) can be modeled as a black-box optimization problem. The goal is to solve $argmin_{x \in \mathcal{X}}f_{obj}(x)$ over an architecture search space $\mathcal{X}$, where $f_{obj}(x)$ is the objective performance metric (e.g., classification error on the validation set) corresponding to the architecture configuration $x$. 
In the following, we first introduce the framework of Bayesian optimization and then the zero-cost proxies used in our proposed framework.

\subsection{Bayesian Optimization}
To solve black-box optimization problems with expensive evaluation costs, Bayesian optimization (BO) follows the framework of sequential model-based optimization. 
A typical BO iteration loops over the following three steps:
1) BO fits a probabilistic surrogate model $M$ based on the observations $D=\{(x_1, y_1),...,(x_{n-1}, y_{n-1})\}$, in which $x_i$ is the configuration evaluated in the $i$-th iteration and $y_i$ is its corresponding observed performance; 2) BO uses the surrogate $M$ to select the most promising configuration $x_n$ by maximizing 
$x_{n}=\arg\max_{x \in \mathcal{X}}a(x; M)$, where $a(x; M)$ is the acquisition function designed to balance the trade-off between exploration and exploitation; 3) BO evaluates the configuration $x_n$ to obtain $y_n$(i.e., train the architecture and obtain its validation performance), and augment the observations $D=D\cup\{(x_n,y_n)\}$.

We adopt the Probabilistic Random Forest~\cite{hutter2011sequential} as the surrogate model and the Expected Improvement (EI)~\cite{jones1998efficient} as the acquisition function. 
The EI is defined as:
\begin{equation}
\small
\label{eq_ei}
a(x; M)=\int_{-\infty}^{\infty} \max(y_{best}-y, 0)p_{M}(y|x)dy,
\end{equation}
where $p_{M}(y|x)$ is the conditional probability of $y$ given $x$ under the surrogate model $M$, and $y_{best}$ is the best observed performance in observations $D$, i.e., $y_{best}=\min\{y_1,..., y_n\}$. Note that, EI only takes the improvement over the best performance into consideration.


\subsection{Zero-cost Proxy}
\label{zp}
Different from the low-cost proxies~\cite{zhou2020econas} which require less training resources, zero-cost proxies are a type of proxies that can be computed at initialization. 
The initial design goal of zero-cost proxies is to better direct exploration in existing NAS algorithms without the expensive training costs. 
Recent researches~\cite{abdelfattah2020zero,krishnakumar2022bench} show that a certain combination of proxies work better than single ones but proxies perform quite differently across different tasks. 
To ensure safe use of proxies, we consider multiple proxies so that at least one is likely to be helpful.
In the following, we describe three metrics with their properties used in our proposed method.

\texttt{snip}~\cite{lee2018snip} is a saliency metric that approximates the loss change when a certain parameter is removed.
\texttt{synflow}~\cite{tanaka2020pruning} optimizes \texttt{snip} to avoid layer collapse when performing parameter pruning. 
While \texttt{snip} that requires a batch of data and the original loss function, \texttt{synflow} computes the product of all parameters as its loss function and thus requires no data.
The formulations are as follows, 
\begin{equation}
\scriptsize
\label{snip_synflow}
    \text{\texttt{snip}:}\ S(\theta)=\left|\frac{\partial L_{snip}}{\partial \theta}\odot \theta\right|, \ \ 
    \text{\texttt{synflow}:}\ 
    S(\theta)=\frac{\partial L_{syn}}{\partial \theta}\odot \theta,
\end{equation}
where $L_{snip}$ is the loss function of a network, $L_{syn}$ is the product of all parameters, and $\odot$ is the Hadamard product.
While both \texttt{snip} and \texttt{synflow} are per-parameter metrics, we extend them to score the entire architecture $x$ following~\cite{abdelfattah2020zero} as $P(x)=-\sum_{\theta \in \Theta}S(\theta)$, where $\Theta$ refers to all the parameters of architecture $x$.

Recent work~\cite{mellor2021neural} introduces a correlation metric \texttt{jacob\_cov} that captures the correlation of activations of different inputs within a network given a batch of data. We refer to the original paper for the detailed derivation of the metric. 
The lower the correlation is, the better the network is expected to be.

The three proxies are selected due to two considerations: 1) All of the three proxies can be computed in a relatively short time using at most a batch of data; 2) The proxies have their own properties, i.e., \texttt{jacob\_cov} highlights the activations while \texttt{snip} and \texttt{synflow} are gradient-based proxies of different inputs.

\section{The Proposed Method}
In this section, we present \sys \ -- our proposed method for efficient Bayesian optimization (BO) with zero-cost proxies. 
In general, \sys alters the sampling procedure of traditional BO to integrate the information from zero-cost proxies, thus generalizing to different underlying BO algorithms. 
To tackle the challenges in the introduction, we will answer the following two questions: 1) how to measure the generalization ability of zero-cost proxies as well as the BO surrogate without prior knowledge, and 2) how to effectively integrate BO with zero-cost proxies during optimization.

\subsection{Generalization Ability Measurement}
The goal of Bayesian optimization (BO) is to iteratively find the best configuration with the optimal objective value. 
In other words, a proxy or a surrogate is helpful if it can order the performance of the given architecture configurations correctly.
Inspired by the concept of some multi-fidelity techniques to measure the fitness of probabilistic surrogates~\cite{li2021mfes},
in our framework, we apply a measurement to dynamically estimate the usefulness of zero-cost proxies and BO surrogates during optimization.
For simplicity, in the following discussion, we assume that the given objective function needs to be minimized. And the smaller the proxy score is, the better the architecture is expected to be.

For zero-cost proxies, we need to measure their ability to fit the ground-truth observations $D$, and thus we apply the Kendall-tau correlation to measure the correctness of their ranking results. To convert the value range to $[0,1]$, we define the metric $G$ as 
$G(P_i;D)=\frac{\tau_{P_i}+1}{2}$.
$\tau_{P_i}$ is the coefficient of the pair set $\{(P_i(x),y)|(x,y) \in D\}$, 
where $P_i(x)$ is the output of the zero-cost proxy $i$ given the architecture configuration $x$, and $y$ is its corresponding ground-truth performance.

For BO surrogate, since it is directly trained on the observations $D$, the above definition only calculates the in-sample error of the BO surrogate and cannot correctly reflect the generalization ability of the surrogate on unseen data points. Therefore, we apply the $k$-fold cross-validation strategy for the BO surrogate. Denote $f$ as the mapping function that maps an observed configuration $x$ to its corresponding fold index as $f(x)$. 
The measurement for the BO surrogate is calculated as,
$G(M;D)=\frac{\tau_M+1}{2}$.
$\tau_M$ is the coefficient of the pair set $\{(M_{-f(x)}(x),y)|(x,y) \in D\}$, 
where $M_{-f(x)}$ refers to the predictive mean of the surrogate trained on observations $D$ with the $f(x)$-th fold left out. 
In this way, $x$ is not used when generating $M_{-f(x)}$, thus the new definition is able to measure the generalization ability of BO surrogate only by using the observations $D$. 
Through this measurement, \sys is able to judge the surrogate during optimization.

Note that, to measure the generalization ability, the Kendall-tau is more appropriate than other intuitive alternatives such as mean squared error or log-likelihood because we do not care about the actual values of the predictions during the optimization. Instead, the framework only needs to identify the location of the optimum.

\subsection{Dynamic Influence Combination}
The original BO selects the next configuration to evaluate by maximizing its acquisition function.
However, the BO surrogate is under-fitted when given few observations, i.e., it can not precisely predict the performance of unseen configurations to guide the selection of configurations. 
Rather than a machine learning model, the zero-cost proxies are formulated metrics, among which some may perform better than the under-fitted surrogate at the beginning of the optimization. On the other hand, as the number of observations grows over time, the generalization ability of BO surrogate gradually outperforms the proxies. In this case, more attention should be paid to the surrogate when selecting the next configuration to evaluate. Therefore, the dynamic influence of the proxies and the surrogate on configuration selection should be considered, and the design is non-trivial.

\begin{algorithm}[tb]
  \small
  \caption{Pseudo code for \textit{Sample} in \sys}
  \label{algo:sample}
  \begin{algorithmic}[1]
  \REQUIRE the observations $D$, the current number of iteration $T$, the number of sampled configurations $Q$, the Bayesian optimization surrogate $M$, the zero-cost proxies $P_{1:K}$, and the temperature hyper-parameter $\tau_0$.
  \ENSURE the next architecture configuration to evaluate.
  \STATE \textbf{if} \emph{$|D| < 5$}, then \textbf{return} a random configuration.
  \STATE compute $G(\cdot;D)$ for each proxy and the surrogate.
  \STATE compute $I(\cdot;D)$ according to Eq.~\ref{eq:influence}.
  \STATE draw $Q$ configurations via random and local sampling.
  \STATE compute EI based on surrogate $M$ according to Eq.~\ref{eq_ei}, and the proxy values for $P_{1:K}$ for each sampled configuration.
  \STATE rank the $Q$ configurations and obtain the ranking value of configuration $x_j$ as $R_M(x_j)$ and $R_{P_i}(x_j)$ for the $i$-th proxy.
  \STATE calculate the combined ranking $CR(x_j)$ for each configuration $x_j$ according to Eq.~\ref{eq:cr}.
  \STATE \textbf{return} the configuration with the lowest CR value.
\end{algorithmic}
\end{algorithm}

Concretely in \sys, we alter the acquisition function for selecting configurations by combining the influence of each component. 
Though there are various methods to combine probabilistic outputs of similar scales (e.g., gPOE~\cite{cao2014generalized}), they can not be directly applied in \sys, as we find that the outputs of proxies are deterministic, and they are of significantly different scales with the surrogate output. 
As a result, we propose to use \textbf{the sum of ranking} instead of directly adding the outputs.
During each BO iteration, we sample $Q$ configurations by random sampling and local sampling on well-performed observed configurations. Then we calculate the EI and proxy values for each sampled configuration. Based on these values, we further rank the $Q$ configurations and obtain the ranking value of $x_j$ as $R_M(x_j)$ and $R_{P_i}(x_j)$ for the BO surrogate and proxies, respectively. Finally, the combined ranking value of $x_j$ is defined as:
\begin{equation}
\scriptsize
\label{eq:cr}
    CR(x_j)=I(M;D)R_M(x_j)+\sum_{i=1}^{K}I(P_i;D)R_{P_i}(x_j),
\end{equation}
where $K$ is the number of applied proxies, and $I(\cdot;D)$ is the measured influence in the current iteration based on $G(\cdot;D)$. 
We use a softmax function with temperature to scale the sum of influence to 1 and use the temperature $\tau$ to control the softness of output distribution. The formulation is as follows,
\begin{equation}
\scriptsize
    I(\cdot;D)=\frac{exp(G(\cdot;D)/ \tau)}{\sum exp(G(\cdot;D)/ \tau)},
    \ \tau=\frac{\tau_0}{1+\text{log}\ T},
\label{eq:influence}
\end{equation}
where $\tau_0$ is the only hyper-parameter, and $T$ is the current number of BO iteration. In each iteration, \sys selects the configuration with the lowest combined ranking (CR) to evaluate. 
As there might be other ways to combine BO and proxies, we show that \sys performs better than other intuitive combinations in the following section.

\subsection{Algorithm Summary}
Algorithm~\ref{algo:sample} illustrates the sampling procedure of \sys. It first computes the generalization ability measurements $G$ (Line 2) and converts them to influence $I$ (Line 3). Then, it ranks the sampled configurations (Lines 4-6) and combines the rankings (Line 7). \sys follows a typical BO framework by replacing the original EI-maximizing procedure with the combined ranking-minimizing procedure.


\subsection{Discussions}
\label{sec:discussion}
In this subsection, we discuss the properties of our proposed \sys as follows:

\noindent\textbf{Extension.}
\sys is independent of the choice of BO surrogate and zero-cost proxies, and users can replace those components with state-of-the-art ones in future researches. Moreover, it also supports other orthogonal methods for acceleration, like transfer learning.

\noindent\textbf{Time Complexity.}
The time complexity of each iteration in \sys is $\mathcal{O}(|D|log|D|)$, which is dominated by the cost of fitting a probabilistic random forest surrogate. Note that since the computation cost of proxy scores is constant during each iteration, it is not taken into account.

\noindent\textbf{Overhead Analysis.}
During each iteration, the computation cost of proxy scores is $QT_p$ in which $Q$ is the number of computed configurations, and $T_p$ is the average cost of computation. In practice, $T_p$ can be calculated in seconds, and $Q$ configurations can be parallelly computed even on a single GPU. We set $Q$ to be 500, and the time cost for each iteration is less than a minute. Compared with the actual training cost (hours), this overhead can almost be ignored.

\noindent\textbf{Convergence Discussion.} 
As the number of observations grows, the proxies accumulate misrankings while the surrogate generalizes better. 
As a result, the generalization ability of the BO surrogate will gradually outperform the proxies. 
Meanwhile, the temperature $\tau$ declines when $T$ grows, which sharpens the distribution and leads to the domination of the BO surrogate on influence. 
Finally, \sys puts almost all the weights on the BO surrogate, and the algorithm reverts to standard BO, which enjoys a convergence guarantee~\cite{hutter2011sequential}. 
We provide the analysis of when the combination function reverts to standard BO.

\begin{assumption}{1}
As proxies are not learning models, $G(P_i;D)$ is relatively stable and has an upper bound. Denote the least upper bound among all proxies as $G_u$; As a learning model, we expect that $G(M;D)$ will consistently outperform all proxies after sufficient rounds $T_e$, and then $G(M;D)$ has a lower bound $G_l$, where $G_l > G_u$.
\end{assumption}

\begin{theorem}
After $T_c=\max\{T_e, \exp(\tau_0 * \frac{\log KQ}{G_l - G_u} -1))\}$ rounds where $K$, $Q$ is the number of proxies and sampled configurations, ProxyBO reverts to standard BO.
\end{theorem}
The proofs are provided in the appendix. We will also illustrate this trend of influence in the following section. 

\noindent\textbf{Difference with Previous Methods.}
While \sys shares the same spirit as previous work~\cite{li2021mfes,feurer2018scalable} that auxiliary information is applied to further improve BO using the normalized Kendall-Tau correlation, we discuss the difference between \sys and previous methods as follows:
\textbf{D1}: The first difference is that the auxiliary information is different (i.e., history surrogates in RGPE, low-fidelity surrogates in MFES, and zero-cost proxies in ProxyBO).
\textbf{D2}: The main difference is how we combine the two parts. As proxies do not provide mean and variance outputs, we propose to combine them in acquisition function instead of surrogate as in RGPE and MFES. However, taking NAS-Bench-101 as an example, the challenges are that: 1) The values are of significantly different scales. The synflow values range from $10^{13}$ to $10^{17}$ while the jacob\_cov values are around 272. 2) The distributions are also different. Most synflow values are close to $10^{14}$ while jacob\_cov values are uniformly distributed. We notice that we only aim to choose the architecture with the largest acquisition value. Therefore, we convert the values into ranks so that they can be directly summed up.
\textbf{D3}: We apply softmax with temperature to ensure that ProxyBO reverts to standard BO given sufficient iterations. The theoretical analysis is also provided above.

\section{Experiments and Results}
\label{exp_sec}
To evaluate \sys, we apply it on several public NAS benchmarks. Compared with state-of-the-art baselines, we list three main insights that we will investigate: 1) \sys can dynamically measure the influence of zero-cost proxies during the search process; 2) \sys can effectively integrate those proxies with the BO procedure without prior knowledge. In other words, good proxies greatly boost performance while bad ones influence little; 3) \sys achieves promising results and can significantly accelerate neural architecture search. It reaches similar performance to other methods while spending much less search time.

\subsection{Experimental Setup}
\textbf{Baselines.}
In the main experiment, we compare the proposed method \sys with the following \textbf{fifteen} baselines --- 
\textit{Five regular methods:} 
(1) Random search (RS), 
(2) REINFORCE (RL)~\cite{williams1992simple}, 
(3) Regularized evolutionary algorithm (REA)~\cite{real2019regularized}, 
(4) Bayesian optimization (BO)~\cite{hutter2011sequential},
(5) Binary relation predictor (BRP)~\cite{dudziak2020brp}
---
\textit{Two multi-fidelity methods:}
(6) BOHB~\cite{falkner2018bohb},
(7) MFES~\cite{li2021mfes},
---
\textit{Two weight-sharing methods:}
(8) DARTS-PT~\cite{wang2021rethinking},
(9) ENAS~\cite{pham2018efficient},
---
\textit{Three zero-cost proxies:}
(10) Snip~\cite{lee2018snip},
(11) Synflow~\cite{tanaka2020pruning},
(12) Jacob\_cov~\cite{mellor2021neural},
---
\textit{Three zero-cost proxy-based methods:}
(13) Warm-up BRP~\cite{abdelfattah2020zero}: BRP-NAS with warm-start based on the relative rankings of proxy scores.
(14) A-REA~\cite{mellor2021neural}: REA that evaluates population with the largest proxy scores.
(15) OMNI~\cite{white2021powerful}: BO using the proxy scores as inputs.

\noindent\textbf{Benchmarks.}
To ensure reproducibility as in previous work~\cite{abdelfattah2020zero}, we conduct the experiments on four public NAS benchmarks: NAS-Bench-101~\cite{ying2019bench}, NAS-Bench-201~\cite{dong2019bench}, NAS-Bench-ASR~\cite{mehrotra2020bench}, and NAS-Bench-301~\cite{siems2020bench} with the \textbf{real-world DARTS search space}. Detailed descriptions are provided in the appendix.


We demonstrate the Spearman $\rho$ of proxy scores related to the test results of all models and top-10\% models on each task in Table~\ref{tab:spearman} on five tasks, and note that $\rho$ is the prior knowledge that can not be obtained before optimization. We observe that no proxy dominates the others on all the tasks, and some correlation coefficients are even negative, leading to a number of incorrect rankings.

\begin{table}[t]
    \centering
    \resizebox{0.98\columnwidth}{!}{
    \begin{tabular}{l|c|c|c}
        \toprule
         & \texttt{snip} & \texttt{synflow} & \texttt{jacob\_cov}\\
        \midrule
        NAS-Bench-101 & -0.16 (-0.00) & 0.37 (0.14) & -0.38 (-0.08)\\
        NB2 CIFAR-10 & 0.60 (-0.36) & 0.72 (0.12) & 0.74 (0.15) \\
        NB2 CIFAR-100 & 0.64 (-0.09) & 0.71 (0.42) & 0.76 (0.06) \\
        NB2 ImageNet16-120 & 0.58 (0.13) & 0.70 (0.55) & 0.75 (0.06)\\
        NAS-Bench-ASR & 0.03 (0.13) & 0.41 (-0.01) & -0.36 (0.06)\\
        \bottomrule
    \end{tabular}
    }
    \caption{Spearman $\rho$ of proxies for all (and top-10\%) architectures in NAS spaces.}
    \label{tab:spearman}
\end{table}

\noindent\textbf{Basic Settings.}
In the following subsections, we 
report the average best test error for NAS-Bench-101, 201, 301 and the average best test Phoneme error rate (PER) for NAS-Bench-ASR. The ``average best'' refers to the best-observed performance during optimization, which is non-increasing.
For evaluation-based methods, we set the time budget as 200 times \textbf{the average cost of an entire evaluation} on each task. For zero-cost proxies, we randomly sample 1000 architectures and report the model with the best proxy score. Each weight-sharing method is run until convergence.
To avoid randomness, weight-sharing methods are repeated 5 times, and the other methods are repeated 30 times.
We plot the mean $\pm$ std. results in the following figures. 

\noindent\textbf{Implementation Details.}
We implement \sys based on OpenBox~\cite{li2021openbox}, a toolkit for black-box optimization. 
The other baselines are implemented following their original papers.
While GPWL and BRP contain specific implementations for each benchmark, we only evaluate them on their supported benchmarks based on the open-source version.
Furthermore, we use Pytorch~\cite{paszke2019pytorch} to train neural networks and calculate proxy scores.
The population size for REA and A-REA is 20;
in BRP and Warm-up BRP, 30 models are sampled to train the predictor during each iteration;
in Warm-up BRP, we randomly sample 256 models to perform a warm start; if not mentioned, the zero-cost proxy-based methods apply \textbf{all} three proxies. 
The $\eta$ and $R$ are set to 3 and 27 in BOHB and MFES, respectively.
The $\tau_0$ in \sys is set to 0.05.
In the last experiment using NAS-Bench-301, we use ``xgb'' as the performance predictor and ``lgb\_runtime'' as the runtime predictor. 
The experiments are conducted on a machine with 64 `AMD EPYC 7702P' CPU cores and two `RTX 2080Ti' GPUs.

\subsection{Empirical Analysis}

\begin{figure}
  \begin{center}
	\includegraphics[width=0.8\linewidth]{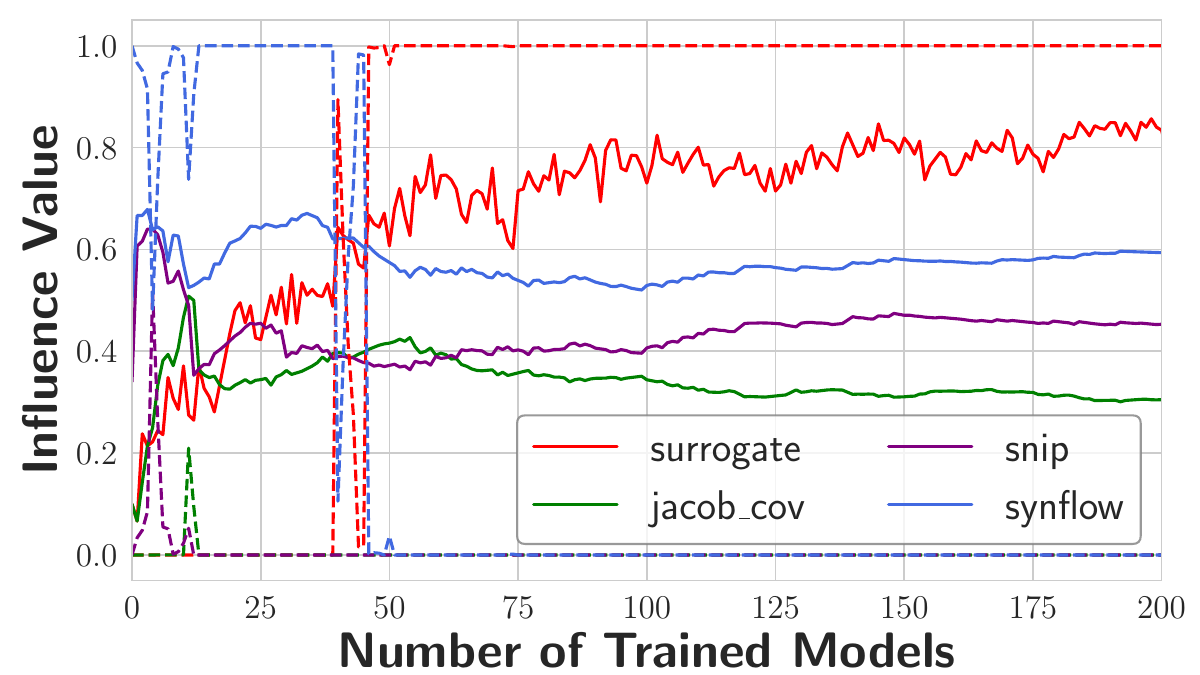}
  \end{center}
  \caption{The solid and dash lines refer to generalization ability measurements and influence values.}
	\label{fig:weight}
\end{figure}

\textbf{\sys can measure the influence of proxies during the search process.}
To show the influence of different proxies, we demonstrate the trend of generalization ability measurements $G$ (solid lines) and influence values $I$ (dash lines) on NAS-Bench-101 in Figure~\ref{fig:weight}. 
As shown in Table~\ref{tab:spearman}, the correlation coefficients of \texttt{snip} and \texttt{jacob\_cov} are negative.
Therefore, their measurements are much lower than \texttt{synflow}'s, and their influence values are kept as zero after about 13 evaluations.
\texttt{synflow}, the best among the three proxies, shows the largest influence in the beginning 36 evaluations.
Our result is also consistent with previous work ~\cite{ning2021evaluating} that using the best proxy performs better than a voting ensemble.
However, since zero-cost proxies are formulated metrics, their measurements are relatively stable. 
The measurement of BO surrogate exceeds \texttt{synflow} at the 39-th evaluation, and it takes the surrogate another 10 evaluations to enlarge the gap.
After that, the generalization ability of the surrogate further increases with more evaluations, and its influence keeps closely to 1. 
In this case, \sys turns back to standard BO with a convergence guarantee.

\begin{table}[tb]

\centering
\setlength{\tabcolsep}{2mm}{
\subtable[NAS-Bench-101 (Optimal: 5.68\%)]{
\resizebox{0.39\pdfpagewidth}{!}{
\begin{tabular}{cccccc}
    \toprule
    Method & None & snip & jacob\_cov & synflow & all \\
    \midrule
    A-REA & 6.19 & 6.27 & 6.62 & 5.95 & 6.07\\
    OMNI (PRF) & 6.14 & 6.28 & 6.37 & 5.91 & 6.05\\
    Warm-up BO (PRF) & 6.14 & 6.31 & 6.30 & 5.91 & 6.08\\
    \sys(PRF) & 6.14 & \textbf{6.18} & \textbf{6.14} & \textbf{5.87} & \textbf{5.96} \\
    \bottomrule
    \end{tabular}
    }
}
}
\centering
\setlength{\tabcolsep}{2mm}{
\subtable[NB2 ImageNet-16-120 (Optimal: 52.69\%)]
{
\resizebox{0.39\pdfpagewidth}{!}{
\begin{tabular}{cccccc}
    \toprule
    Method & None & snip & jacob\_cov & synflow & all \\
    \midrule
    A-REA & 53.08 & 53.43 & 53.15 & 52.80 & 52.91 \\
    OMNI (PRF) & 53.12 & 53.21 & 53.07 & 52.83 & 52.92\\
    Warm-up BO (PRF) & 53.12 & 53.24 & 53.20 & 52.77 & 53.08 \\
    \sys(PRF) & 53.12 & \textbf{53.18} & \textbf{52.93} & \textbf{52.74} & \textbf{52.82} \\
    \bottomrule
    \end{tabular}
    }
}
}
\caption {Mean test errors (\%) of zero-cost proxy-based methods with different proxies.}
\label{table_zero}
\end{table}

\begin{table*}[tbh]
    
    \centering
    \resizebox{0.82\pdfpagewidth}{!}{
    \begin{tabular}{ccccccc}
    \toprule
    Method & Runtime (\#Eval) & {NAS-Bench-101} & {NB2-CIFAR-10} & {NB2-CIFAR-100} & {NB2-ImageNet16-120} & {NAS-Bench-ASR} \\
    \midrule
    \multicolumn{7}{c}{\textbf{Regular Methods}}\\
    RS & 200 & $6.34\pm0.12$ & $9.11\pm0.21$ & $27.97\pm0.66$ & $54.01\pm0.55$ & $21.61\pm0.10$ \\
    RL & 200 & $6.31\pm0.14$ & $9.02\pm0.24$ & $27.66\pm0.65$ & $53.58\pm0.45$ & $21.62\pm0.09$\\
    REA & 200 & $6.19\pm0.24$ & $8.62\pm0.21$ & $26.67\pm0.35$ & $53.08\pm0.36$ & $21.50\pm0.07$ \\
    BO & 200 & $6.14\pm0.23$ & $8.80\pm0.22$ & $27.03\pm0.45$ & $53.12\pm0.37$ & $21.47\pm0.06$ \\
    BRP & 200 & $6.05\pm0.16$ & $8.58\pm0.13$ & $26.58\pm0.12$ & $52.96\pm0.29$ & $21.50\pm0.08$ \\
    \midrule
    \multicolumn{7}{c}{\textbf{Multi-fidelity Methods}}\\
    BOHB & 200 & $6.26\pm0.18$ & $8.95\pm0.28$ & $27.65\pm0.72$ & $53.77\pm0.53$ & $21.74\pm0.16$\\
    MFES & 200 & $6.10\pm0.17$ & $8.71\pm0.18$ & $26.57\pm0.13$ & $53.19\pm0.18$ & $21.71\pm0.12$\\
    \midrule
    \multicolumn{7}{c}{\textbf{Weight-sharing Methods}}\\
    ENAS & $\approx$7 & $8.17\pm0.42$ & $46.11\pm0.58$ & $86.04\pm2.33$ & $85.19\pm2.10$ & $24.45\pm0.90$\\
    DARTS-PT & $\approx$18 & $7.79\pm0.61$ & $15.33\pm2.23$ & $34.03\pm2.24$ & $61.36\pm1.91$ & $24.08\pm0.43$  \\
    \midrule
    \multicolumn{7}{c}{\textbf{Zero-cost Proxies}}\\
    Snip & $<$1 & $10.68\pm2.16$ & $13.45\pm1.80$ & $36.41\pm3.36$ & $71.94\pm9.09$ & $31.61\pm18.17$\\
    Jacob\_cov & $<$1 & $13.86\pm1.86$ & $12.19\pm1.60$ & $32.99\pm2.84$ & $60.43\pm4.46$ & $69.95\pm24.67$\\
    Synflow & $<$1 & $8.32\pm1.64$ & $10.30\pm0.94$ & $29.55\pm1.77$ & $56.94\pm3.57$ & $25.70\pm12.91$\\
    \midrule
    \multicolumn{7}{c}{\textbf{Zero-cost Proxy-based Methods}}\\
    A-REA & 200 & $6.07\pm0.21$ & \bm{$8.54\pm0.08$} & $26.59\pm0.11$ & $52.91\pm0.24$ & $21.47\pm0.06$\\
    Warm-up BRP & 200 & $6.04\pm0.15$ & $8.58\pm0.21$ & $26.60\pm0.31$ & $53.02\pm0.35$ & $21.51\pm0.10$\\
    OMNI & 200 & $6.05\pm0.11$ & $8.60\pm0.14$ & $26.64\pm0.29$ & $52.92\pm0.26$ & $21.48\pm0.05$ \\
    \sys & 200 & $\bm{5.96\pm0.13}$ & $\bm{8.54\pm0.10}$ & $\bm{26.52\pm0.17}$ & $\bm{52.82\pm0.19}$ & $\bm{21.43\pm0.03}$ \\
    \midrule
    Optimal & / & $5.68$ & $8.48$ & $26.49$ & $52.69$ & $21.40$ \\
    \bottomrule
    \end{tabular}
    }
    \caption{Mean $\pm$ std. test errors (\%) on NAS-Bench-101 and NAS-Bench-201, and test PERs (\%) on NAS-Bench-ASR. ``NB2'' refers to NAS-Bench-201, and ``Optimal'' refers to the ground-truth optima. The evaluation number of multi-fidelity methods, weight-sharing methods, and zero-cost proxies is computed by their runtime divided by the average training time of architectures. }
    \label{tab:e2e_results}
\end{table*}

\noindent\textbf{\sys can effectively integrate BO procedure with zero-cost proxies.}
To show that combining BO and proxies are non-trivial, we compare different methods of using zero-proxy proxies.
Based on probabilistic random forest (PRF), we add the baselines OMNI~\cite{white2021powerful} and Warm-up BO with 40 start points of good proxy scores.
While BO (PRF) can not be directly extended to support moving proposal, we use A-REA as the baseline.
The results of using different proxies are presented in Table~\ref{table_zero}.
We find that:
1) \sys outperforms Warm-up and OMNI. The reason is that, 
Warm-up applies a static setting rather than the dynamic one in \sys, and the different distribution of proxy scores across tasks makes it difficult to rescale the scores as inputs for OMNI.
2) 
If the correlations are not accessible, \sys performs the best among baselines using all proxies.
When using bad proxies with negative correlation (see Table~\ref{tab:spearman}), \sys 
performs similar to that without proxies.
If we know the best correlated proxy (synflow) beforehand, \sys saves budget for identifying bad proxies. 
It reduces the regret (i.e., the distance to the global optima) of the second-best method by \textbf{17-38\%} and performs better than using all proxies. 
We provide additional ablation study, effectiveness analysis, and discussions on how to select $\tau_0$ based on prior knowledge in the appendix.

\noindent\textbf{\sys achieves promising results.} 
Table~\ref{tab:e2e_results} shows the test results on five tasks given the budget of 200 evaluations. 
We observe that zero-cost proxies require extremely short computation time, but the final results are not satisfactory. 
The reason is that it can not correctly rank the most-accurate architectures (see top-10\% architectures in Table~\ref{tab:spearman}). 
Note that, though weight-sharing methods and zero-cost proxies require less budget than evaluation-based methods, their performance has converged. 
In addition, since the multi-fidelity methods conduct a large number of evaluations with fewer epochs, they train fewer models to convergence than the standard BO, which may lead to an under-estimation of converged model performance given limited budget. 
In our experiments, we observe that MFES performs worse than BO on NB2-ImageNet16-120 and NAS-Bench-ASR.
Finally, due to the limitations of the warm-up strategy, Warm-up BRP slightly outperforms BRP on NAS-Bench-101 but performs worse on the other benchmarks.
While A-REA can not distinguish bad proxies, it performs well on NB2 where each proxy is effective but less competitive on other benchmarks.
Among the competitive baselines, \sys achieves the best average test results on all five tasks and remarkablely, it reduces the test regret of the best baseline by \textbf{70\%} on NB2-CIFAR-100.

We also apply \sys to state-of-the-art surrogate GPWL~\cite{ru2020interpretable} on NAS-Bench-201 and NAS-Bench-101.
The corresponding results are shown in Table~\ref{tab:sota}.
All methods use the same strategy to sample candidates for fair comparison.
We find that \sys(GPWL) achieves the best results on four tasks.
Since using GPWL alone is empirically superior to PRF, 
the gain of applying \sys is smaller on GPWL than on PRF.
But it still decreases the regret of BO (GPWL) by \textbf{25-84\%} on four tasks, which indicates the effectiveness of \sys on utilizing useful zero-cost proxies.

In addition, we compare \sys with SOTA weight-sharing method $\beta$-DARTS~\cite{ye2022b} on NB2-ImageNet16-120. \sys may not obtain a satisfactory result as fast as weight-sharing methods due to initialization. It takes $\beta$-DARTS and \sys 15 and 28 GPU hours to obtain the same error (53.66\%). However, when given a larger budget, the error of \sys continuously decreases while $\beta$-DARTS has already converged. The error decrease of \sys over $\beta$-DARTS given different times of budget are shown in Table~\ref{tab:gap_beta_darts}. While the results of \sys (16x, 52.94\%) approach the optimum (52.69\%), this improvement is relatively considerable. Users can decide whether to use \sys and how much budget to offer by balancing the tradeoff between time and performance. More detailed discussion and results are provided in the appendix.

\begin{table}[tb]
   
    \centering
    \resizebox{0.28\pdfpagewidth}{!}{
    \begin{tabular}{ccccccc}
    \toprule
      & 1.87x & 4x & 8x & 16x & 32x\\
    \midrule
    Gap & 0.00 & 0.22 & 0.53 & 0.72 & 0.85\\
    \bottomrule
    \end{tabular}
    }
     \caption{Error decrease (\%) when given more time budget compared with $\beta$-DARTS on NB2-ImageNet16-120.}
    \label{tab:gap_beta_darts}
\end{table}

\begin{figure*}[tb]
	\centering
	\subfigure[NB201 CIFAR-10]{
		\scalebox{0.31}{
			\includegraphics[width=1\linewidth]{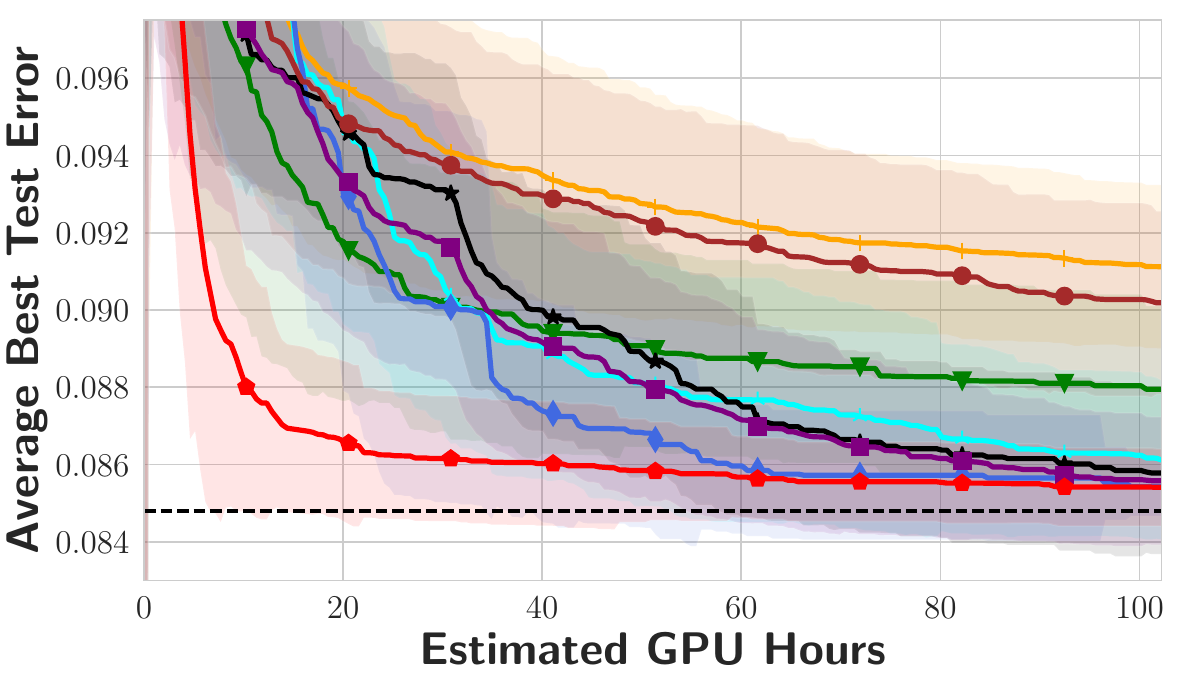}
	}}
	\subfigure[NB201 CIFAR-100]{
		\scalebox{0.31}{
			\includegraphics[width=1\linewidth]{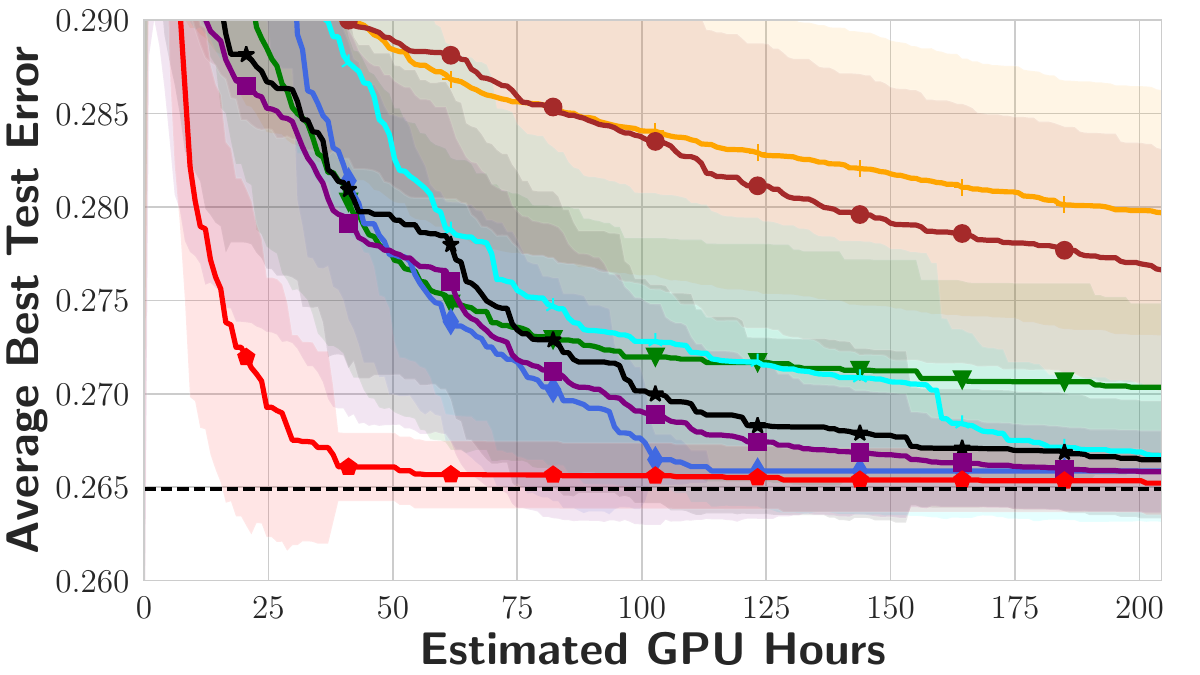}
	}}
	\subfigure[NB201 ImageNet16-120]{
		\scalebox{0.31}{
			\includegraphics[width=1\linewidth]{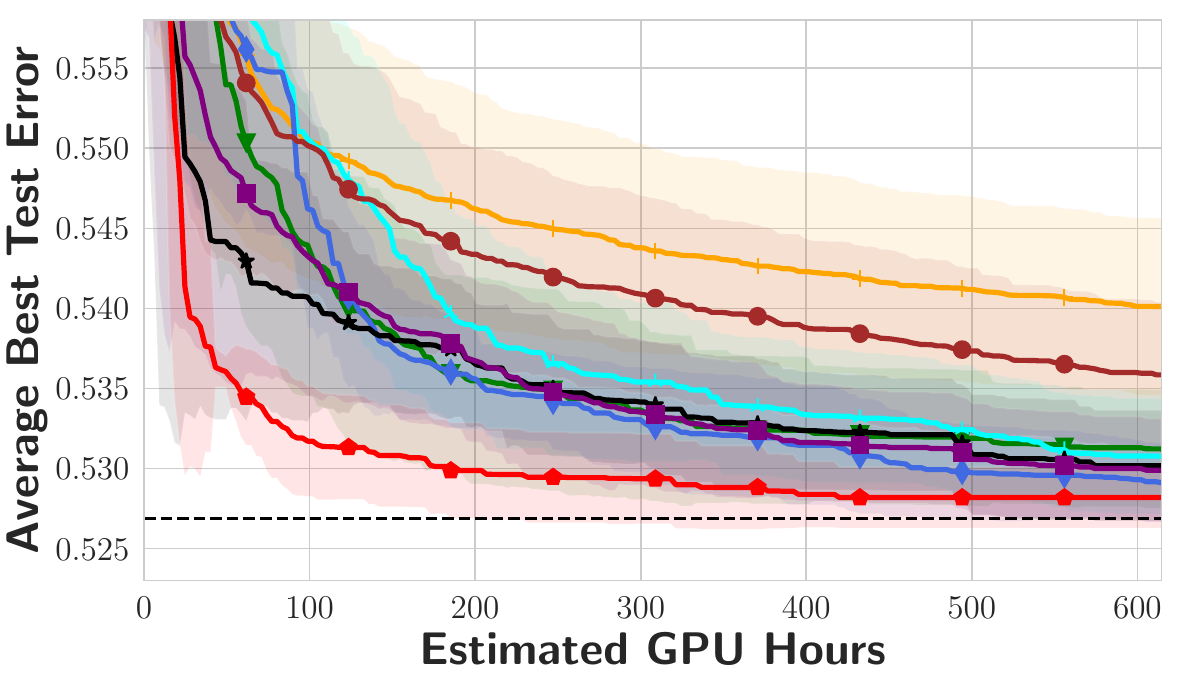}
	}}
	\subfigure[NAS-Bench-101]{
		\scalebox{0.31}{
			\includegraphics[width=1\linewidth]{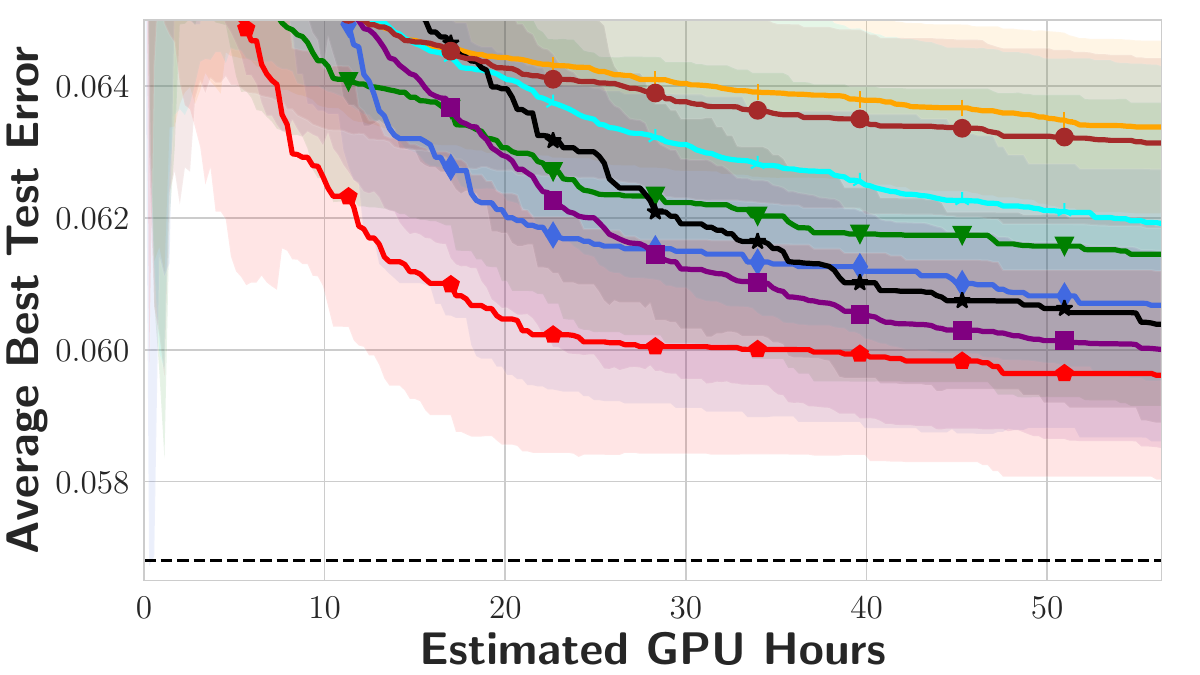}
	}}
	\subfigure[NAS-Bench-ASR]{
		\scalebox{0.31}{
			\includegraphics[width=1\linewidth]{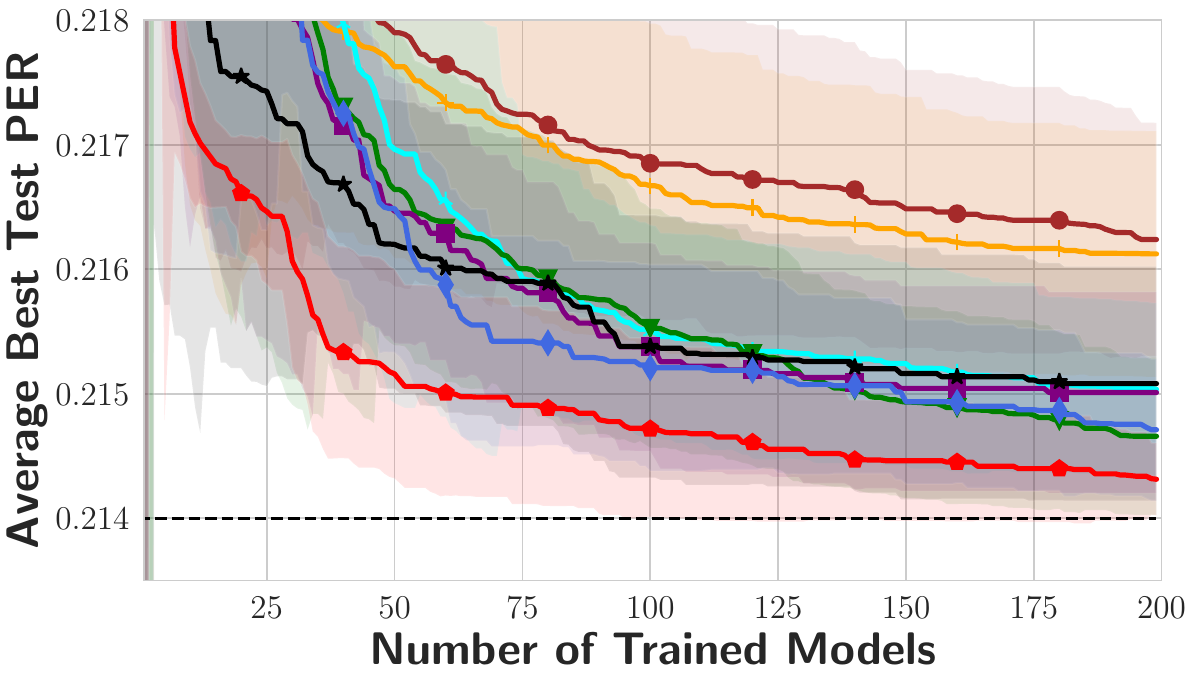}
	}}
	\subfigure{
		\scalebox{0.31}{			\raisebox{0.75\height}{\includegraphics[width=0.95\linewidth]{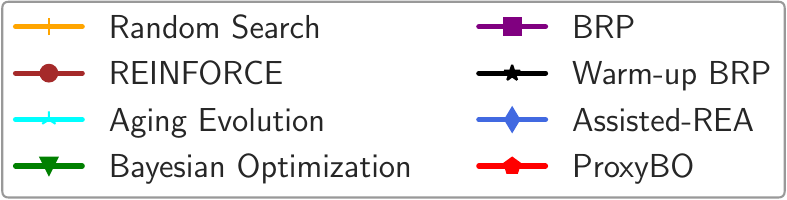}}
	}}
	\caption{Test results during neural architecture search on three benchmarks. The black dash lines refer to the global optima.}
  \label{fig:end2end}
\end{figure*}

\begin{table}[tpb!]
  \centering
  
  \resizebox{0.39\pdfpagewidth}{!}{
\begin{tabular}{ccccc}
    \toprule
    Method & NB101 & NB2 C10 & NB2 C100 & NB2 I16  \\
    \midrule
    BO (PRF) & 6.14 & 8.80 & 27.03 & 53.12\\
    \sys(PRF) & 5.96 & 8.54 & 26.52 & 52.82 \\
    \midrule
    BO (GPWL) & 5.88 & 8.56 & 26.55 & 52.97\\
    \sys(GPWL) & \textbf{5.83} & \textbf{8.52} & \textbf{26.50} & \textbf{52.78}\\
    \hline
    Optimal & 5.68 & 8.48 & 26.49 & 52.69\\
    \bottomrule
    \end{tabular}
}
\captionof{table}{Mean test errors (\%) with different surrogates.}
\label{tab:sota}
\end{table}

\begin{table}[tpb!]
  \centering
  
  \resizebox{0.37\pdfpagewidth}{!}{
    \begin{tabular}{ccccc}
    \toprule
    & BRP & W-BRP & A-REA & \sys \\
    \midrule
    NAS-Bench-101 & 94 & 105 & 74 & \textbf{41} \\
    NB2 CIFAR-10 & 170 & 178 & 108 & \textbf{44}  \\
    NB2 CIFAR-100 & 142 & 184 & 98 & \textbf{37} \\
    NB2 ImageNet16-120 & 144 & 168 & 144 & \textbf{46} \\
    NAS-Bench-ASR & 179 & 213 & 177 &\textbf{51}\\
    \bottomrule
    \end{tabular}
    }
    \captionof{table}{Number of evaluations required to achieve the same average results as REA with 200 evaluations.}
  \label{tab:speedup}
\end{table}

\noindent\textbf{\sys can significantly accelerate NAS.}
Figure~\ref{fig:end2end} demonstrates the search results of regular and zero-cost proxy-based methods.
Consistent with Table~\ref{tab:e2e_results}, we observe that Warm-up BRP only shows acceleration in early rounds on NB2 ImageNet16-120 and NAS-Bench-ASR.
Compared with weight-sharing methods and zero-cost proxies, it takes \sys less than 8 evaluations to surpass their converged results. 
In addition, the test results of \sys decrease rapidly before 75 iterations, and it consistently outperforms the other baselines on the five tasks.
To show the speedup of \sys, we further compare the number of trained models required to achieve the same average results as REA using 200 evaluations in Table~\ref{tab:speedup}. 
Concretely, \sys achieves $\bm{3.92-5.41\times}$ and $\bm{2.29-3.86\times}$ speedups relative to the state-of-the-art method REA and BRP, respectively.

We also evaluate \sys on realistic DARTS search space using NAS-Bench-301. 
While BANANAS~\cite{white2021bananas} is a state-of-the-art BO method on that space, we apply \sys to BANANAS and compare it with evaluation-based methods, and the results are shown in Table~\ref{tab:speedup_301}.
As previous study~\cite{white2021powerful} claims that local search is not so effective on large space given limited budget, REA requires about twice the budget as BANANAS.
Among compared methods, \sys outperforms A-REA, and it spends $\bm{0.70\times}$ GPU hours to achieve the same results as BANANAS. 
More detailed results on NAS-Bench-301 are provided in the appendix.

In addition, we evaluate BANANAS and ProxyBO (BANANAS) on DARTS space using CIFAR10. The search takes nearly 3 days, and the final architecture is trained given 600 epochs. 
To save evaluation costs during searching, we train 8-cell architectures instead of 20-cell ones for 40 epochs. 
The final test errors of BANANAS and ProxyBO are 2.75\% and 2.68\%, respectively, which show that ProxyBO can further improve BANANAS.

\begin{table}[tb]
    
    \centering
    \resizebox{0.38\pdfpagewidth}{!}{
    \begin{tabular}{cccccc}
    \toprule
      & RS & REA & A-REA & BANANAS & \sys\\
    \midrule
    \#Eval & $\approx$2k & 441 & 304 & 200 & \textbf{142}\\
    GPU Hours & $\approx$3k & 663 & 424 & 289 & \textbf{201}\\
    \bottomrule
    \end{tabular}
    }
    \caption{Number of evaluations required to achieve the same results as BANANAS (5.10\%).}
    \label{tab:speedup_301}
\end{table}


\section{Conclusion}
\label{Conclusion}
In this paper, we introduced \sys, an efficient Bayesian optimization framework that leverages the auxiliary knowledge from zero-cost proxies. In \sys, we proposed two components, namely the generalization ability measurement and dynamic influence combination, which tackles the unforeseeable reliability and one-shot usage issues in existing methods, and developed a more principled way to utilize zero-cost proxies.
We evaluated \sys on four public benchmarks and demonstrated its superiority over competitive baselines.


\section{Acknowledgements}
This work is supported by NSFC (No. 61832001 and U22B2037) and Kuaishou-PKU joint program. Wentao Zhang and Bin Cui are the corresponding authors. Special thanks to Tianyi Bai and Yupeng Lu for their help during the rebuttal period.

\nocite{abraham2021fairlof,wei2022robust,zhu2021pre,zhang2020snapshot,zhang2022pasca,yang2022diffusion,li2022volcanoml}

\bibliography{reference}

\begin{thebibliography}{54}
\providecommand{\natexlab}[1]{#1}

\bibitem[{Abdelfattah et~al.(2021)Abdelfattah, Mehrotra, Dudziak, and
  Lane}]{abdelfattah2020zero}
Abdelfattah, M.~S.; Mehrotra, A.; Dudziak, {\L}.; and Lane, N.~D. 2021.
\newblock Zero-Cost Proxies for Lightweight NAS.
\newblock In \emph{International Conference on Learning Representations}.

\bibitem[{Abraham et~al.(2021)}]{abraham2021fairlof}
Abraham, S.~S.; et~al. 2021.
\newblock FairLOF: Fairness in Outlier Detection.
\newblock \emph{Data Science and Engineering}, 6(4): 485--499.

\bibitem[{Cai et~al.(2018)Cai, Chen, Zhang, Yu, and Wang}]{cai2018efficient}
Cai, H.; Chen, T.; Zhang, W.; Yu, Y.; and Wang, J. 2018.
\newblock Efficient architecture search by network transformation.
\newblock In \emph{Proceedings of the AAAI Conference on Artificial
  Intelligence}, volume~32.

\bibitem[{Cao and Fleet(2014)}]{cao2014generalized}
Cao, Y.; and Fleet, D.~J. 2014.
\newblock Generalized product of experts for automatic and principled fusion of
  Gaussian process predictions.
\newblock \emph{arXiv preprint arXiv:1410.7827}.

\bibitem[{Chen, Gong, and Wang(2021)}]{chen2020neural}
Chen, W.; Gong, X.; and Wang, Z. 2021.
\newblock Neural Architecture Search on ImageNet in Four GPU Hours: A
  Theoretically Inspired Perspective.
\newblock In \emph{International Conference on Learning Representations}.

\bibitem[{Dong and Yang(2020)}]{dong2019bench}
Dong, X.; and Yang, Y. 2020.
\newblock NAS-Bench-201: Extending the Scope of Reproducible Neural
  Architecture Search.
\newblock In \emph{International Conference on Learning Representations}.

\bibitem[{Dudziak et~al.(2020)Dudziak, Chau, Abdelfattah, Lee, Kim, and
  Lane}]{dudziak2020brp}
Dudziak, L.; Chau, T.; Abdelfattah, M.; Lee, R.; Kim, H.; and Lane, N. 2020.
\newblock BRP-NAS: Prediction-based NAS using GCNs.
\newblock \emph{Advances in Neural Information Processing Systems}, 33.

\bibitem[{Falkner, Klein, and Hutter(2018)}]{falkner2018bohb}
Falkner, S.; Klein, A.; and Hutter, F. 2018.
\newblock BOHB: Robust and efficient hyperparameter optimization at scale.
\newblock In \emph{International Conference on Machine Learning}, 1437--1446.
  PMLR.

\bibitem[{Feurer, Letham, and Bakshy(2018)}]{feurer2018scalable}
Feurer, M.; Letham, B.; and Bakshy, E. 2018.
\newblock Scalable meta-learning for bayesian optimization using
  ranking-weighted gaussian process ensembles.
\newblock In \emph{AutoML Workshop at ICML}.

\bibitem[{He et~al.(2016)He, Zhang, Ren, and Sun}]{he2016deep}
He, K.; Zhang, X.; Ren, S.; and Sun, J. 2016.
\newblock Deep residual learning for image recognition.
\newblock In \emph{Proceedings of the IEEE conference on computer vision and
  pattern recognition}, 770--778.

\bibitem[{Huang et~al.(2017)Huang, Liu, Van Der~Maaten, and
  Weinberger}]{huang2017densely}
Huang, G.; Liu, Z.; Van Der~Maaten, L.; and Weinberger, K.~Q. 2017.
\newblock Densely connected convolutional networks.
\newblock In \emph{Proceedings of the IEEE conference on computer vision and
  pattern recognition}, 4700--4708.

\bibitem[{Hutter, Hoos, and Leyton-Brown(2011)}]{hutter2011sequential}
Hutter, F.; Hoos, H.~H.; and Leyton-Brown, K. 2011.
\newblock Sequential model-based optimization for general algorithm
  configuration.
\newblock In \emph{International Conference on Learning and Intelligent
  Optimization}, 507--523. Springer.

\bibitem[{Jones, Schonlau, and Welch(1998)}]{jones1998efficient}
Jones, D.~R.; Schonlau, M.; and Welch, W.~J. 1998.
\newblock Efficient global optimization of expensive black-box functions.
\newblock \emph{Journal of Global optimization}, 13(4): 455--492.

\bibitem[{Kandasamy et~al.(2018)Kandasamy, Neiswanger, Schneider, P{\'o}czos,
  and Xing}]{kandasamy2018neural}
Kandasamy, K.; Neiswanger, W.; Schneider, J.; P{\'o}czos, B.; and Xing, E.~P.
  2018.
\newblock Neural architecture search with Bayesian optimisation and optimal
  transport.
\newblock In \emph{Proceedings of the 32nd International Conference on Neural
  Information Processing Systems}, 2020--2029.

\bibitem[{Krishnakumar et~al.(2022)Krishnakumar, White, Zela, Tu, Safari, and
  Hutter}]{krishnakumar2022bench}
Krishnakumar, A.; White, C.; Zela, A.; Tu, R.; Safari, M.; and Hutter, F. 2022.
\newblock NAS-Bench-Suite-Zero: Accelerating Research on Zero Cost Proxies.
\newblock \emph{arXiv preprint arXiv:2210.03230}.

\bibitem[{Lee, Hyung, and Hwang(2021)}]{lee2021rapid}
Lee, H.; Hyung, E.; and Hwang, S.~J. 2021.
\newblock Rapid neural architecture search by learning to generate graphs from
  datasets.
\newblock \emph{arXiv preprint arXiv:2107.00860}.

\bibitem[{Lee, Ajanthan, and Torr(2018)}]{lee2018snip}
Lee, N.; Ajanthan, T.; and Torr, P. 2018.
\newblock Snip: Single-shot Network Pruning Based on Connection Sensitivity.
\newblock In \emph{International Conference on Learning Representations}.

\bibitem[{Li et~al.(2017)Li, Jamieson, DeSalvo, Rostamizadeh, and
  Talwalkar}]{li2017hyperband}
Li, L.; Jamieson, K.; DeSalvo, G.; Rostamizadeh, A.; and Talwalkar, A. 2017.
\newblock Hyperband: A novel bandit-based approach to hyperparameter
  optimization.
\newblock \emph{The Journal of Machine Learning Research}, 18(1): 6765--6816.

\bibitem[{Li et~al.(2022{\natexlab{a}})Li, Shen, Jiang, Zhang, Li, Liu, Zhang,
  and Cui}]{li2022hyper}
Li, Y.; Shen, Y.; Jiang, H.; Zhang, W.; Li, J.; Liu, J.; Zhang, C.; and Cui, B.
  2022{\natexlab{a}}.
\newblock Hyper-Tune: Towards Efficient Hyper-parameter Tuning at Scale.
\newblock \emph{arXiv preprint arXiv:2201.06834}.

\bibitem[{Li et~al.(2021{\natexlab{a}})Li, Shen, Jiang, Gao, Zhang, and
  Cui}]{li2021mfes}
Li, Y.; Shen, Y.; Jiang, J.; Gao, J.; Zhang, C.; and Cui, B.
  2021{\natexlab{a}}.
\newblock MFES-HB: Efficient Hyperband with Multi-Fidelity Quality
  Measurements.
\newblock In \emph{Proceedings of the AAAI Conference on Artificial
  Intelligence}, volume~35, 8491--8500.

\bibitem[{Li et~al.(2021{\natexlab{b}})Li, Shen, Zhang, Chen, Jiang, Liu,
  Jiang, Gao, Wu, Yang, Zhang, and Cui}]{li2021openbox}
Li, Y.; Shen, Y.; Zhang, W.; Chen, Y.; Jiang, H.; Liu, M.; Jiang, J.; Gao, J.;
  Wu, W.; Yang, Z.; Zhang, C.; and Cui, B. 2021{\natexlab{b}}.
\newblock OpenBox: A Generalized Black-box Optimization Service.
\newblock \emph{Proceedings of the 27th ACM SIGKDD Conference on Knowledge
  Discovery \& Data Mining}.

\bibitem[{Li et~al.(2022{\natexlab{b}})Li, Shen, Zhang, Zhang, and
  Cui}]{li2022volcanoml}
Li, Y.; Shen, Y.; Zhang, W.; Zhang, C.; and Cui, B. 2022{\natexlab{b}}.
\newblock VolcanoML: speeding up end-to-end AutoML via scalable search space
  decomposition.
\newblock \emph{The VLDB Journal}, 1--25.

\bibitem[{Liu, Simonyan, and Yang(2019)}]{liu2018darts}
Liu, H.; Simonyan, K.; and Yang, Y. 2019.
\newblock DARTS: Differentiable Architecture Search.
\newblock In \emph{International Conference on Learning Representations}.

\bibitem[{Ma, Cui, and Yang(2019)}]{ma2019deep}
Ma, L.; Cui, J.; and Yang, B. 2019.
\newblock Deep neural architecture search with deep graph bayesian
  optimization.
\newblock In \emph{2019 IEEE/WIC/ACM International Conference on Web
  Intelligence (WI)}, 500--507. IEEE.

\bibitem[{Mehrotra et~al.(2021)Mehrotra, Ramos, Bhattacharya, Dudziak,
  Vipperla, Chau, Abdelfattah, Ishtiaq, and Lane}]{mehrotra2020bench}
Mehrotra, A.; Ramos, A. G.~C.; Bhattacharya, S.; Dudziak, {\L}.; Vipperla, R.;
  Chau, T.; Abdelfattah, M.~S.; Ishtiaq, S.; and Lane, N.~D. 2021.
\newblock NAS-Bench-ASR: Reproducible Neural Architecture Search for Speech
  Recognition.
\newblock In \emph{International Conference on Learning Representations}.

\bibitem[{Mellor et~al.(2021)Mellor, Turner, Storkey, and
  Crowley}]{mellor2021neural}
Mellor, J.; Turner, J.; Storkey, A.; and Crowley, E.~J. 2021.
\newblock Neural architecture search without training.
\newblock In \emph{International Conference on Machine Learning}, 7588--7598.
  PMLR.

\bibitem[{Ning et~al.(2021)Ning, Tang, Li, Zhou, Liang, Yang, and
  Wang}]{ning2021evaluating}
Ning, X.; Tang, C.; Li, W.; Zhou, Z.; Liang, S.; Yang, H.; and Wang, Y. 2021.
\newblock Evaluating Efficient Performance Estimators of Neural Architectures.
\newblock \emph{Advances in Neural Information Processing Systems}, 34.

\bibitem[{Paszke et~al.(2019)Paszke, Gross, Massa, Lerer, Bradbury, Chanan,
  Killeen, Lin, Gimelshein, Antiga et~al.}]{paszke2019pytorch}
Paszke, A.; Gross, S.; Massa, F.; Lerer, A.; Bradbury, J.; Chanan, G.; Killeen,
  T.; Lin, Z.; Gimelshein, N.; Antiga, L.; et~al. 2019.
\newblock Pytorch: An imperative style, high-performance deep learning library.
\newblock \emph{Advances in neural information processing systems}, 32:
  8026--8037.

\bibitem[{Pham et~al.(2018)Pham, Guan, Zoph, Le, and Dean}]{pham2018efficient}
Pham, H.; Guan, M.; Zoph, B.; Le, Q.; and Dean, J. 2018.
\newblock Efficient neural architecture search via parameters sharing.
\newblock In \emph{International Conference on Machine Learning}, 4095--4104.
  PMLR.

\bibitem[{Real et~al.(2019)Real, Aggarwal, Huang, and Le}]{real2019regularized}
Real, E.; Aggarwal, A.; Huang, Y.; and Le, Q.~V. 2019.
\newblock Regularized evolution for image classifier architecture search.
\newblock In \emph{Proceedings of the AAAI conference on artificial
  intelligence}, volume~33, 4780--4789.

\bibitem[{Ru et~al.(2021)Ru, Wan, Dong, and Osborne}]{ru2020interpretable}
Ru, B.; Wan, X.; Dong, X.; and Osborne, M. 2021.
\newblock Interpretable Neural Architecture Search via Bayesian Optimisation
  with Weisfeiler-Lehman Kernels.
\newblock In \emph{International Conference on Learning Representations}.

\bibitem[{Shu et~al.(2021)Shu, Cai, Dai, Ooi, and Low}]{shu2021nasi}
Shu, Y.; Cai, S.; Dai, Z.; Ooi, B.~C.; and Low, B. K.~H. 2021.
\newblock NASI: Label-and Data-agnostic Neural Architecture Search at
  Initialization.
\newblock \emph{arXiv preprint arXiv:2109.00817}.

\bibitem[{Shu et~al.(2022)Shu, Dai, Wu, and Low}]{shu2022unifying}
Shu, Y.; Dai, Z.; Wu, Z.; and Low, B. K.~H. 2022.
\newblock Unifying and Boosting Gradient-Based Training-Free Neural
  Architecture Search.
\newblock \emph{arXiv preprint arXiv:2201.09785}.

\bibitem[{Siems et~al.(2020)Siems, Zimmer, Zela, Lukasik, Keuper, and
  Hutter}]{siems2020bench}
Siems, J.; Zimmer, L.; Zela, A.; Lukasik, J.; Keuper, M.; and Hutter, F. 2020.
\newblock NAS-Bench-301 and the case for surrogate benchmarks for neural
  architecture search.
\newblock \emph{arXiv preprint arXiv:2008.09777}.

\bibitem[{Snoek, Larochelle, and Adams(2012)}]{snoek2012practical}
Snoek, J.; Larochelle, H.; and Adams, R.~P. 2012.
\newblock Practical bayesian optimization of machine learning algorithms.
\newblock In \emph{Advances in neural information processing systems},
  2951--2959.

\bibitem[{So, Le, and Liang(2019)}]{so2019evolved}
So, D.; Le, Q.; and Liang, C. 2019.
\newblock The evolved transformer.
\newblock In \emph{International Conference on Machine Learning}, 5877--5886.
  PMLR.

\bibitem[{Springenberg et~al.(2016)Springenberg, Klein, Falkner, and
  Hutter}]{springenberg2016bayesian}
Springenberg, J.~T.; Klein, A.; Falkner, S.; and Hutter, F. 2016.
\newblock Bayesian optimization with robust Bayesian neural networks.
\newblock \emph{Advances in neural information processing systems}, 29:
  4134--4142.

\bibitem[{Tanaka et~al.(2020)Tanaka, Kunin, Yamins, and
  Ganguli}]{tanaka2020pruning}
Tanaka, H.; Kunin, D.; Yamins, D.~L.; and Ganguli, S. 2020.
\newblock Pruning neural networks without any data by iteratively conserving
  synaptic flow.
\newblock \emph{Advances in Neural Information Processing Systems}, 33.

\bibitem[{Theis et~al.(2018)Theis, Korshunova, Tejani, and
  Husz{\'a}r}]{theis2018faster}
Theis, L.; Korshunova, I.; Tejani, A.; and Husz{\'a}r, F. 2018.
\newblock Faster gaze prediction with dense networks and fisher pruning.
\newblock \emph{arXiv preprint arXiv:1801.05787}.

\bibitem[{Wang, Zhang, and Grosse(2020)}]{wang2019picking}
Wang, C.; Zhang, G.; and Grosse, R. 2020.
\newblock Picking Winning Tickets Before Training by Preserving Gradient Flow.
\newblock In \emph{International Conference on Learning Representations}.

\bibitem[{Wang et~al.(2021)Wang, Cheng, Chen, Tang, and
  Hsieh}]{wang2021rethinking}
Wang, R.; Cheng, M.; Chen, X.; Tang, X.; and Hsieh, C.-J. 2021.
\newblock Rethinking Architecture Selection in Differentiable NAS.
\newblock In \emph{International Conference on Learning Representations}.

\bibitem[{Wei et~al.(2022)Wei, Wang, Tu, and Li}]{wei2022robust}
Wei, T.; Wang, H.; Tu, W.; and Li, Y. 2022.
\newblock Robust model selection for positive and unlabeled learning with
  constraints.
\newblock \emph{Science China Information Sciences}, 65(11): 1--13.

\bibitem[{White, Neiswanger, and Savani(2021)}]{white2021bananas}
White, C.; Neiswanger, W.; and Savani, Y. 2021.
\newblock BANANAS: Bayesian Optimization with Neural Architectures for Neural
  Architecture Search.
\newblock In \emph{Proceedings of the AAAI Conference on Artificial
  Intelligence}, volume~35, 10293--10301.

\bibitem[{White et~al.(2021)White, Zela, Ru, Liu, and
  Hutter}]{white2021powerful}
White, C.; Zela, A.; Ru, B.; Liu, Y.; and Hutter, F. 2021.
\newblock How Powerful are Performance Predictors in Neural Architecture
  Search?
\newblock \emph{arXiv preprint arXiv:2104.01177}.

\bibitem[{Williams(1992)}]{williams1992simple}
Williams, R.~J. 1992.
\newblock Simple statistical gradient-following algorithms for connectionist
  reinforcement learning.
\newblock \emph{Machine learning}, 8(3): 229--256.

\bibitem[{Xu et~al.(2021)Xu, Zhao, Lin, Gao, Sun, and Yang}]{xu2021knas}
Xu, J.; Zhao, L.; Lin, J.; Gao, R.; Sun, X.; and Yang, H. 2021.
\newblock KNAS: green neural architecture search.
\newblock In \emph{International Conference on Machine Learning}, 11613--11625.
  PMLR.

\bibitem[{Yang, Zhang, and Hong(2022)}]{yang2022diffusion}
Yang, L.; Zhang, Z.; and Hong, S. 2022.
\newblock Diffusion Models: A Comprehensive Survey of Methods and Applications.
\newblock \emph{arXiv preprint arXiv:2209.00796}.

\bibitem[{Ye et~al.(2022)Ye, Li, Li, Chen, Fan, and Ouyang}]{ye2022b}
Ye, P.; Li, B.; Li, Y.; Chen, T.; Fan, J.; and Ouyang, W. 2022.
\newblock b-DARTS: Beta-Decay Regularization for Differentiable Architecture
  Search.
\newblock In \emph{Proceedings of the IEEE/CVF Conference on Computer Vision
  and Pattern Recognition}, 10874--10883.

\bibitem[{Ying et~al.(2019)Ying, Klein, Christiansen, Real, Murphy, and
  Hutter}]{ying2019bench}
Ying, C.; Klein, A.; Christiansen, E.; Real, E.; Murphy, K.; and Hutter, F.
  2019.
\newblock Nas-bench-101: Towards reproducible neural architecture search.
\newblock In \emph{International Conference on Machine Learning}, 7105--7114.
  PMLR.

\bibitem[{Zhang et~al.(2020)Zhang, Jiang, Shao, and Cui}]{zhang2020snapshot}
Zhang, W.; Jiang, J.; Shao, Y.; and Cui, B. 2020.
\newblock Snapshot boosting: a fast ensemble framework for deep neural
  networks.
\newblock \emph{Science China Information Sciences}, 63(1): 1--12.

\bibitem[{Zhang et~al.(2022)Zhang, Shen, Lin, Li, Li, Ouyang, Tao, Yang, and
  Cui}]{zhang2022pasca}
Zhang, W.; Shen, Y.; Lin, Z.; Li, Y.; Li, X.; Ouyang, W.; Tao, Y.; Yang, Z.;
  and Cui, B. 2022.
\newblock Pasca: A graph neural architecture search system under the scalable
  paradigm.
\newblock In \emph{Proceedings of the ACM Web Conference 2022}, 1817--1828.

\bibitem[{Zhou et~al.(2020)Zhou, Zhou, Zhang, Loy, Yi, Zhang, and
  Ouyang}]{zhou2020econas}
Zhou, D.; Zhou, X.; Zhang, W.; Loy, C.~C.; Yi, S.; Zhang, X.; and Ouyang, W.
  2020.
\newblock Econas: Finding proxies for economical neural architecture search.
\newblock In \emph{Proceedings of the IEEE/CVF Conference on Computer Vision
  and Pattern Recognition}, 11396--11404.

\bibitem[{Zhu, Dai, and Chen(2021)}]{zhu2021pre}
Zhu, D.-H.; Dai, X.-Y.; and Chen, J.-J. 2021.
\newblock Pre-Train and Learn: Preserving Global Information for Graph Neural
  Networks.
\newblock \emph{Journal of Computer Science and Technology}, 36(6): 1420--1430.

\bibitem[{Zoph et~al.(2018)Zoph, Vasudevan, Shlens, and Le}]{zoph2018learning}
Zoph, B.; Vasudevan, V.; Shlens, J.; and Le, Q.~V. 2018.
\newblock Learning transferable architectures for scalable image recognition.
\newblock In \emph{Proceedings of the IEEE conference on computer vision and
  pattern recognition}, 8697--8710.

\end{thebibliography}

\end{document}